# Spending Money Wisely: Online Electronic Coupon Allocation based on Real-Time User Intent Detection

Liangwei Li*
Liucheng Sun*
leon.llw@alibaba-inc.com
liucheng.slc@alibaba-inc.com
Alibaba Group
Hangzhou, Zhejiang

Chenwei Weng
wengchenwei.pt@alibaba-inc.com
Alibaba Group
Hangzhou, Zhejiang

Chengfu Huo
chengfu.huocf@alibaba-inc.com
Alibaba Group
Hangzhou, Zhejiang

Weijun Ren
afei@alibaba-inc.com
Alibaba Group
Hangzhou, Zhejiang

## ABSTRACT

Online electronic coupon (e-coupon) is becoming a primary tool for e-commerce platforms to attract users to place orders. E-coupons are the digital equivalent of traditional paper coupons which provide customers with discounts or gifts. One of the fundamental problems related is how to deliver e-coupons with minimal cost while users' willingness to place an order is maximized. We call this problem the coupon allocation problem. This is a non-trivial problem since the number of regular users on a mature e-platform often reaches hundreds of millions and the types of e-coupons to be allocated are often multiple. The policy space is extremely large and the online allocation has to satisfy a budget constraint. Besides, one can never observe the responses of one user under different policies which increases the uncertainty of the policy making process. Previous work fails to deal with these challenges. In this paper, we decompose the coupon allocation task into two subtasks: the user intent detection task and the allocation task. Accordingly, we propose a two-stage solution: at the first stage (*detection stage*), we put forward a novel Instantaneous Intent Detection Network (IIDN) which takes the user-coupon features as input and predicts user real-time intents; at the second stage (*allocation stage*), we model the allocation problem as a Multiple-Choice Knapsack Problem (MCKP) and provide a computational efficient allocation method using the intents predicted at the detection stage. Long Short Term Memory (LSTM) and a special attention mechanism are applied on IIDN to better describe temporal dependencies of sequential features. And we manage to solve the imbalanced label problem for the user intent detection task with a brand new perspective by using the logical relationship between multiple user intents. We conduct extensive online and offline experiments and the results show the superiority of our proposed framework, which has brought great profits to the platform and continues to function online.

## 1 INTRODUCTION

In the field of marketing, a coupon is a document provided by a merchant or an organization that promises a financial discount or rebate when purchasing a product. Online electronic coupon (e-coupon) is the digital equivalent of traditional coupons. Due to the increasingly fierce business competition between e-commerce platforms, e-coupon is becoming a primary tool to attract users to place orders. One of the fundamental problems related to e-coupon is how to deliver them with minimal cost while the users' willingness to pay is maximized, which is the coupon allocation problem. The coupon allocation is non-trivial because of two key challenges. (1) The number of regular users on large e-commerce platforms often reaches hundreds of millions and there are multiple types of coupons to be allocated. The policy space is combinatorial and very large. The allocation requires also an online decision under a predefined budget constraint. These challenges make finding the optimal policy extremely hard. (2) For each user, the policy maker can never observe his or her responses under different allocation policies since a user can only be in one state at any time. This increases the uncertainty of the policy making process. A straightforward way to solve this problem is to borrow methods from the resource allocation community. In recent year, there has been an increasing interest in this field [3, 8, 10, 19, 31]. However, they fail to deal with the challenges mentioned above.

In this work, we put forward a novel framework IIDN-MCKP to overcome these obstacles. We do not make decisions directly on the raw policy space. Instead, we decompose the coupon allocation problem into two subtasks: the user intent detection task and the allocation task. And we propose a two-stage framework accordingly.

The two key challenges can be settled with our proposed two-stage framework. The reason is two-fold. First, the two-stage decomposition is a popular choice to address large, combinatorial action space problems [28, 33]. Each subtask is relatively simpler to solve than the original task and the original task can be further solved by combining the results of the subtasks. Second, by detecting the user purchasing intent, we can estimate the the potential promotion of the coupon and make decisions accordingly. However, the intent detection is not so simple. As the online e-commerce environment is complex, identifying precisely a user's intent is also challenging. Various factors will affect the user's intent. For instance, a user is possibly not willing to place an order at first. But a glance at a product may stimulate his or her desire to buy and the intent will change accordingly. A good intent detector must capture these weak signals. Besides, the label imbalance for the purchasing intent detection is also problematic. Not all the users place orders on the e-commerce platform and thus the distribution of samples across the classes can be highly skewed. We will detail solutions to these problems in the following parts.

At the first stage, which we call the detection stage, we identify the user purchasing intent (with or without coupons) using a novel deep neural network called Instantaneous Intent Detection Network (IIDN). To make the intent detection more precise, we collect two types of user features: the real-time features and the static features. The real-time features are user real-time behavior sequences gathered from the edge devices and the static features are history information cached on the cloud server. IIDN automatically extract useful information from the raw input without heavy feature engineering. In particular, IIDN uses Long Short Term Memory (LSTM) to model the temporal dependencies of the real-time sequence. An attention mechanism is applied to fuse the outputs of the LSTM layer. An encoder encapsulates all the input features and passes the encoded information to the decoder. At the decoder, we introduce an auxiliary staying intent detection task to overcome the imbalanced label problem of the purchasing intent detection. A standard Recurrent Neural Network (RNN) decoding process then goes on to estimate the desired intents.

At the second stage, which we call the allocation stage, we use the intents predicted at the detection stage and we model the allocation problem as a Multiple-Choice Knapsack Problem (MCKP). Specifically, we try to optimize a global object function under a series of constraints including the budget constraint. We adopt a well-studied primal-dual framework to solve the optimization problem.

The contribution of this work is three-fold. (1) We formulate the online conpon allocation problem and propose a novel two-stage framework to approach this hard problem. (2) We put forward a strong user intent detector called IIDN which automatically learns from the raw input data. IIDN also solves the imbalanced label problem by introducing an auxiliary task. (3) We conduct extensive experiments to evaluation the performances of our proposed framework IIDN-MCKP. Both offline and online experiments show the superiority of IIDN-MCKP. The proposed allocation system has brought great profits to the platform and still functions online.

## 2 RELATED WORK
### 2.1 Resource Allocation

Coupon allocation is a resource allocation (RA) problem. The RA is the process of assigning and managing assets in a way that an organization's goals are achieved. Ahuja et al. [3] develops optimization methods to solve problems related to wireless network, crowdsourcing systems, and healthcare systems. Csáji et al. [8] and Gai et al. [10] formulate the RA problem as a Markov Decision Process (MDP) and apply the approximate dynamic programming methods. Pesavento et al. [19] and Wang et al. [31] employ machine learning methods to solve the network optimization problems.

In our settings, the coupon allocation problem has two key features. First this is typically an online, multi-dimensional packing problem where the policy space is extremely large, which contributes to its hardness. Second, for each user, one can never observe simultaneously the responses under different coupon allocation policies. Users have different sensitiveness to coupons. We propose to analyze user intents before the actual coupon allocation.

### 2.2 User Intent Detection

Detecting user intent refers to predicting which intent corresponds best to a user request. The problem of user intent detection has been heavily studied in the machine learning community. In [16], authors present a semantically enriched word embedding method to improve intent detection task. In [35], Capsule Neural Networks are used to detecting emerging user intents where no labeled utterances are currently available. Wang et al. [32] propose a bi-model based recurrent network to perform the intent detection and slot filling tasks jointly. With the rapid development of e-commerce, user purchasing intent detection on e-commerce platforms is also a promising research field. Basically, given a user-click session, the goal of purchasing intent detection is to predict whether the user is going to place an order. A session is usually formed by a sequence of user behaviors. [24, 29, 34] adopt the recurrent neural network to model the sequence nature of sessions. [11] creatively collect touch-interactive features to better describe the current user.

In this work, the purchasing intent is also the key task. However, previous work fails to deal with the imbalanced label challenge in the purchasing intent prediction task: only a small fraction of users buy something on the platform. One solution is to apply random or synthetic undersampling/oversampling [6, 7, 25, 27]. But the sampling method changes the original distribution of labels. Others reweight the training examples and improve the training objective [15, 20]. However, the reweighting coefficients are hard to determine when the problem becomes complicated. Besides, the objective functions need to be carefully designed in these methods. Herein, we propose a novel solution where we introduce an auxiliary task called the staying intent detection task. As any user leaves the platform at last, the label imbalance issue doesn't exist for the staying intent. This auxiliary task provides information gain for the purchasing intent, which will be detailed in Section 3.1.

Besides, we make full use of the edge computing [4] to collect instaneous user features. These real-time features are combined with the static user features using a novel deep neural network to improve the intent detection. We also deploy the deep learning model on edge devices to reduce the communication cost between the cloud server and the devices.

## 2.3 Recurrent Models

Recurrent Neural Network (RNN) is an augmented feed forward network containing the edges that span adjacent time steps. This introduces a notion of time to the model. At time $t$, nodes with recurrent edges receive the input $x^{(t)}$ and the last hidden state $q^{(t-1)}$. The RNN produces the output $y^{(t)}$ and the next hidden state $q^{(t)}$. It's proven that the vanilla RNN suffers from the vanishing gradient problem. Long Short Term Memory (LSTM) [12] is introduced to overcome this problem. LSTM resembles the standard RNN with a hidden layer, but each ordinary node in the hidden layer is replaced by a memory cell which ensures that the gradient passes across many time steps without vanishing or exploding.

## 2.4 Multiple-Choice Knapsack Problem

The knapsack problem is a classic problem in operations research and theoretical computer science and has a wide range of applications [14]. Specifically, in our task, we are faced with an online multiple-choice knapsack problem (MCKP), where we need to make an instant and irrevocable decision when a user arrives. The corresponding online algorithm is well studied under different assumptions [18]. In this paper, we take the stochastic assumption where the users are drawn from a stationary distribution and the adopted method falls into the primal-dual framework [2].

## 3 METHODOLOGY

In this section, we first formulate the problem of coupon allocation, and then present the overall structure of our proposed IIDN-MCKP. IIDN-MCKP is a two-stage framework: at detection stage, we identify the real-time user intents using Instantaneous Intent Detection Network (IIDN); at the second stage, we model the allocation problem as a MCKP and we provide the allocation strategy based on the intents detected at the detection stage.

### 3.1 Preliminaries

Let $\mathcal{U} = \{u\}$ be the set of all users. The goal of coupon allocation is to find an optimal allocation policy

$$\begin{aligned}\pi^* &= \arg\max_{\pi} R(\mathcal{U}, \pi) \\ s.t. \quad & C(\mathcal{U}, \pi) \leq B\end{aligned} \quad (1)$$

where $\pi : \mathcal{U} \to \mathbb{R}^+$ denotes the allocation policy, $R(\mathcal{U}, \pi)$ denotes the conversion rate (i.e., proportion of buyers) of the user group $\mathcal{U}$ under the policy $\pi$, $C(\mathcal{U}, \pi)$ denotes the cost under the policy $\pi$ and $B$ denotes the budget constraint. A reasonable allocation policy should make sure the coupons are delivered to the most coupon-sensitive users to get a good return on investment (ROI). Hence, the real-time user intent must be detected precisely before making any allocation strategy. The whole decision process is thus divided into two stages: the detection stage and the allocation stage.

At the detection stage, each user $u_i$ is represented by a tuple of user features $(s, h, c)$ where $s$ stands for real-time user features (item clicked, visit time, etc.), $h$ stands for static user features (user level, age level, etc.) and $c$ stands for the coupon amount (0 included). We emphasize the distinction between $s$ and $h$ as they are heterogenous.

In our scenario, the key intent to detect is the purchasing intent $pP(s, h, c) = p(pay|s, h, c)$. However, as mentioned in Section 2.2, the imbalanced label problem for the purchasing intent presents in our case. To solve this problem, we introduce an auxiliary task: detecting the staying intent $pS(s, h, c) = p(stay|s, h, c)$. There are two considerations behind this decision. (1) Staying on the platform can be viewed as a necessary condition for any other intent. A user needs to "be willing to stay" in order to "be willing to buy". Mathematically:

$$p(pay, stay|s, h, c) = p(pay|s, h, c) \quad (2)$$

And the following equation holds if Equation 2 holds:

$$\underbrace{p(pay|s, h, c)}_{pP(s, h, c)} = \underbrace{p(stay|s, h, c)}_{pS(s, h, c)} \times \underbrace{p(pay|stay, s, h, c)}_{pPS(s, h, c)} \quad (3)$$

The imbalanced label problem of estimating $pP(s, h, c)$ is caused by the fact that not all users place orders on the platform, whereas any user leaves the platform at last. The imbalanced label problem does not exist in estimating the staying intent. By estimating first the auxiliary $pS(s, h, c)$, $pP(s, h, c)$ can be estimated more accurately since more clues besides the purchasing information are utilized to assist the purchasing intent estimation. The $pP(s, h, c)$ estimation enjoys an information gain brought by $pPS(s, h, c)$. (2) The staying intent can also be used at the allocation stage. Coupons are only delivered to uses who are more likely to leave, which avoids delivering coupons to users who are still attracted by the platform.

At the allocation stage, we model the whole problem as a MCKP. For each coupon $j$, let $c_j$ denote its amount. For simplicity, available coupon amounts are assumed to be the same for all users and are known in advance. The total value of coupons delivered can not exceed a predetermined budget $B$. $v_{ij}$ and $s_{ij}$ are defined as:

$$\begin{aligned} v_{ij} &= pP(s, h, c_j) \\ s_{ij} &= pS(s, h, c_j) \end{aligned} \quad (4)$$

where $i$ refers to the index of the user $u_i \in \mathcal{U}$. We define a binary decision variable $x_{ij}$ that equals to 1 if and only if coupon $j$ is delivered to user $u_i$. We define further a staying interest threshold $\gamma$ which is used to filter users who are leaving. The objective is to maximize the total conversion rate of $\mathcal{U}$ under the budget constraint. Using the notations above, for all the users such that $s_{ij} \leq \gamma$, we solve the following optimization problem:

$$\begin{aligned} \max & \sum_{i=1}^{M} \sum_{j=1}^{N} v_{ij} x_{ij} \\ s.t. & \sum_{i=1}^{M} \sum_{j=1}^{N} c_j x_{ij} \leq B, \\ & \sum_{j}^{N} x_{ij} \leq 1, \quad \forall i \\ & x_{ij} \geq 0, \quad \forall i, j \end{aligned} \quad (5)$$

### 3.2 Instantaneous Intent Detection Network

Figure 1 depicts the full architecture of IIDN. On the whole, IIDN is fed with a sequence of real-time user features $s = (s_1, s_2, ..., s_n)$ (tap, visit time, etc.), the static user features $h$ (age, vip level, etc.) and the amount $c$ of coupon whose influence on the current user is to be estimated; it then predicts the purchasing probability $pP(s, h, c)$ and the staying probability $pS(s, h, c)$. IIDN contains an embedding lookup layer which maps $s$ and $h$ to dense vectors $s_d = (s_{d,1}, s_{d,2}, ..., s_{d,n})$ and $h_d$. A recurrent layer using LSTM is applied to $s_d$ to model the temporal dependencies. Then an attention layer fuses the outputs from the recurrent layer and produces a feature map $f$ describing the real-time characteristics of the current user. An encoder processes $(f, h_d, c)$ and builds a fixed-length vector representation $v$. Conditioned on the encoded representation, a decoder using RNN, ReLU and softmax estimates $pP(s, h, c)$ and $pS(s, h, c)$.

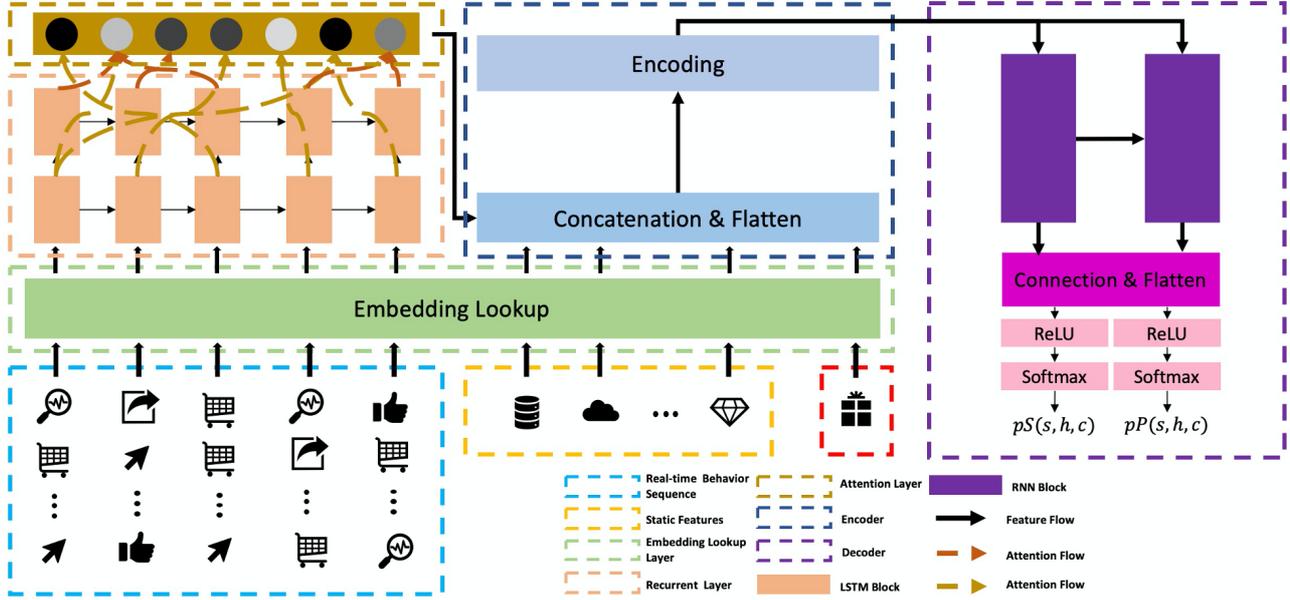

Figure 1: Model Architecture of IIDN.

#### 3.2.1 Embedding Lookup Layer.
The raw values of $s$, $h$ and $c$ are highly heterogenous, which contain discrete tap features, continuous duration features, high dimensional one-hot category features, etc. This heterogeneity prevents the deep network from extracting valid features. The embedding lookup layer embeds $s$, $h$ and $c$ into scaled dense vectors $s_d$, $h_d$ and $c_d$. We adopt the table lookup mechanism from [11] to transform these raw features into low dimensional dense representations. For the simplicity of notation, we will use $s_t$, $h$ and $c$ to denote $s_{d,t}$, $h_d$ and $c_d$ in the follow parts of this paper.

#### 3.2.2 LSTM Layer.
The sequential nature of the real-time user features inspires us to use RNN to model the temporal dependencies between these features. While vanilla RNN works in principle, it would be hard to train the RNN due to the long-term dependencies [5]. And in our setting, feature sequences may be extremely long (more than 100). Thus vanilla RNN fails to guarantee a stable learning process. The LSTM is proven to learn long range temporal dependencies, so LSTM is used here to deal with the input sequences with variable length. The application of LSTM can be summarized concisely as:

$$\begin{aligned}
g_t &= \sigma(W_g \otimes [q_{t-1}, s_t] + b_g) \\
i_t &= \sigma(W_i \otimes [q_{t-1}, s_t] + b_i) \\
\tilde{c}_t &= tanh(W_c \odot [q_{t-1}, s_t] + b_c) \\
c_t &= g_t \odot c_{t-1} + i_t \odot \tilde{c}_t \\
o_t &= \sigma(W_o \otimes [q_{t-1}, s_t] + b_o) \\
q_t &= o_t \odot tanh(c_t)
\end{aligned} \quad (6)$$

where $\otimes$ denotes the matrix product operator, $\odot$ denotes the element-wise product operator, $s_t$ is the embedded sequence at the $t$-th time step, $q_t$ is the $t$-th hidden state and $c_t$ is the $t$-th cell state.

#### 3.2.3 Attention Layer.
Attention mechanism is known to be able to automatically discover global dependencies between input and output [30]. We apply a slightly different attention mechanism. Instead of using merely the final output state of the LSTM layers, we apply attention on every output state of the LSTM layers to capture the semantics about the user intent at all scales. The attention layers are lightweight so their overhead of computational cost is negligible.

#### 3.2.4 Encoder.
Each user is characterized by a tuple of features $(s, h, c)$. While the real-time features $s$ provide essential clues to the current intents, the static information such as age and vip level are also important. Thus the encoder consumes the concatenation of the feature map $f$, the static information $h$ and the coupon amount $c$ and produce a fixed length vector representation containing necessary information to infer user intents. We use Multilayer Perceptron (MLP) to build the encoder.

#### 3.2.5 Decoder.
The decoder predicts multiple user intents by decoding the encoded information from the encoder. As mentioned in Section 3.1, we estimate the purchasing intent $pP(s, h, c)$ by first estimating the staying intent $pS(s, h, c)$ to avoid the imbalanced label issue. As shown in Equation 3, the purchasing intent and the staying intent are connected by a conditional probability which can be viewed as an information gain in the purchasing intent estimation task. Hence we propose a RNN-based decoder to fully explore this single-directional dependency. The decoding process follows a standard RNN decoding formulation [26].

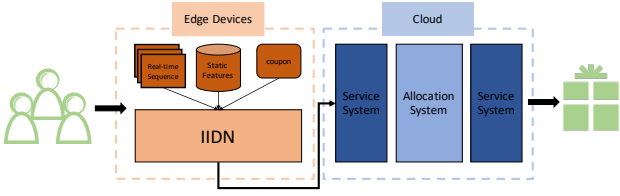

Figure 2: Overall structure of the Coupon Allocation System.

*3.2.6  Loss Design.* The loss functions of the staying intent and the purchasing intent are defined as follows:

$$\mathcal{L}_s = -\frac{1}{N}\sum_{(x,y_s)}^{N}(y_s log pS(x) + (1-y_s)log(1-pS(x)))$$
$$\mathcal{L}_p = -\frac{1}{N}\sum_{(x,y_p)}^{N}(y_p log pP(x) + (1-y_p)log(1-pP(x)))$$
(7)

where $x = (s, h, c)$ is the feature tuple, $y_s$ is the label indicating whether a user keeps staying, $y_p$ is the label indicating whether a user places an order and $N$ is the size of the training set. The goal of IIDN is to minimize the global loss function:

$$\mathcal{L} = \mathcal{L}_s + \mathcal{L}_p \qquad (8)$$

## 3.3  MCKP-Allocation

We adopt the primal-dual framework proposed by [2] to solve the problem defined in Equation 5. Let $\alpha$ and $\beta_j$ be the associated dual variables respectively. After obtaining the dual variables, we can solve the problem in an online fashion. Precisely, according to the principle of the primal-dual framework, we have the following allocation rule:

$$x_{ij} = \begin{cases} 1, & \text{where } j = \arg\max_i(v_{ij} - \alpha c_j) \\ 0, & \text{otherwise} \end{cases} \qquad (9)$$

## 4  SYSTEM DEPLOYMENT

Benefit from the fast development of mobile devices, we are able to collect real-time features to improve the performance of intent detection. Besides, the strict latency constraints (less than 10ms) required by the e-commerce mobile applications makes it impossible to deploy the IIDN on the server side following the conventional cloud-based architecture. Otherwise, the high latency caused by the data transmission from the edge devices to the server is unacceptable in these real world applications. Recent advances in edge computing provides us with an alternative way to overcome this problem. We propose to deploy the IIDN on the mobile devices and the server only accomplishes the allocation task.

Table 1: Statistics of the features used.

| Feature Type | Feature Name | Num. of Features |
|---|---|---|
| Real-time Features | Num. of Collections Num. of favorites ... Page Staying Duration | 15 |
| Static Features | Age Level VIP Level ... Category Preference | 89 |
| Coupon Feature | Coupon Amount | 1 |

Figure 2 depicts the overall structure of our design of the coupon allocation system. We have deployed the system on a large-scale e-commerce platform *1688*[1] and it serves more than 5,000,000 users during Alibaba 0331 promotion. The system still functions online. The procedure is as follow. When a user launches the application, the user's static information is sent to the mobile device and cached in the storage system. Each time the allocation process is triggered, the real-time features and the static features are concatenated with the coupon statistics to feed into the IIDN. IIDN then predicts the two probabilities $pP(s, h, c)$ and $pS(s, h, c)$. These two scores are transferred to the server to finish the allocation phase. Note that the two floating point scores are lightweight and the communication cost is negligible.

## 5  EXPERIMENTS

In this section, we present the experimental setup and the corresponding results for our two-stage settings. For the first stage, we estimate the detection accuracy of real-time user intents; for the second stage, we demonstrate the coupon allocation performance.

### 5.1  Experimental Setup

*5.1.1  Experimental Settings.* In accordance with our two-stage settings, the whole experiment consists of two parts: the intent detection evaluation part and the coupon allocation evaluation part. For the intent detection part, we mainly conduct the experiment on a large-scale industrial dataset collected from *1688*. We collect 1,200,000 users' data (under the privacy code) for two weeks. For each user, we collect their real-time features obtained from edge devices (smart phones, tablet PCs, etc.) and their static information cached on the server. The real-time features are highly serialized which contain user activities on the APP at different time steps. The static information mainly contains user registration statistics and history records. Coupon amount is also used as a feature. The statistics of the features used in our experiments are shown in Table 1. For the allocation part, we conduct the experiment in the real world environment where more than 5,000,000 users are involved. The rest of the experimental settings remain the same as the first part.

*5.1.2  Compared Methods.* In this section, we discuss the compared methods. For the intent detection part, we compare three baseline models with our proposed IIDN. Besides, we validate the effect of each component in IIDN. All the methods hold the same embedding look up layer.

---
[1] www.1688.com

- **LR** [23] is the Logistic Regression which is a strong baseline in many binary classification problems. Sequential real-time features and static features are simply concatenated together to form the input of the LR model.
- **GBDT** [21] is a competitive gradient boosting model widely used in industrial environment. The features are processed similarly with the LR model.
- **RNN+DNN** [33] is used to model the temporal dependencies of the real-time features and to verify the usage of LSTM.
- **IIDN-single-LSTM** is IIDN with a single LSTM layer.
- **IIDN-non-attention** is IIDN with double LSTM layers but without the attention layer.
- **IIDN-non-auxiliary-task** is IIDN with double LSTM layers but without the auxiliary staying intent prediction task.
- **IIDN** is our proposed intent detection method with a double-LSTM layer, an attention layer and the auxiliary staying intent prediction task.

For the allocation part, we verify the MCKP modeling. We manually build the feature tuple $\{(s, h, c_i)\}_i$ where $\{c_i\}_i$ represents the available amounts. The feature tuple is fed into IIDN and we obtain the purchasing intents $\{pP(s, h, c_i)\}_i$ under different coupon amounts. We then compare the following allocation strategies:

- **Non-allocation** where no one gets the coupon.
- **All-allocation** where every one gets the coupon with the amount randomly selected from the available amounts.
- **Uplift-allocation** [22] where the amount to be allocated is

$$c^* = \arg\max_i pP(s, h, c_i) - pP(s, h, c_0)$$
$$s.t. pP(s, h, c_i) - pP(s, h, c_0) \geq \alpha, \forall i \quad (10)$$

where $pP(s, h, c_0)$ denotes the purchasing intent without coupon and $\alpha$ denotes the minimum uplift score for any coupon amount. If no amount reaches the threshold, then no coupon is allocated to the current user.
- **IIDN-MCKP** is our proposed two-stage method described in Section 3.

*5.1.3 Training Details.* IIDN is trained with SGD using the Adam Optimizer [17] with the hyper-parameters of $\epsilon = 0.0001$, $\beta_1 = 0.9$, $\beta_2 = 0.999$. The maximum sequence length of the real-time features is 100. The output dimension of the embedding look up layer is 256. The dimension of the hidden state of the LSTM is 128. We train IIDN using a distributed Tensorflow [1] with 1 parameter server and 50 workers.

## 5.2 Intent Detection Evaluation

*5.2.1 Metrics.* At this stage, we focus on the intent prediction performance where Area Under ROC Curve (AUC) [9] and Logloss [13] are widely used. AUC measures the pairwise ranking performance of the classification results and Logloss is used to measure the event probability prediction. We mainly finish the evaluation on the large-scale industrial dataset mentioned above.

*5.2.2 Results.* The intent detection results are shown in Table 2. We have the following findings. First, in terms of AUC and Logloss,

Table 2: Results of different intent detection methods.

| Method | AUC | Logloss |
| --- | --- | --- |
| LR | 0.644 | 0.358 |
| GBDT | 0.658 | 0.342 |
| RNN+DNN | 0.692 | 0.168 |
| IIDN-single-LSTM | 0.703 | 0.155 |
| IIDN-non-attention | 0.712 | 0.149 |
| IIDN-non-auxiliary-task | 0.707 | 0.156 |
| IIDN | **0.738** | **0.127** |

IIDN outperforms the baseline methods LR and GBDT. This validates the sequential modeling of temporal dependencies in the real-time user features. Second, RNN+DNN is inferior to the LSTM-based models. This phenomenon verifies the adoption of LSTM in IIDN. We also find that the double-layer LSTM outperforms its single-layer counterpart. This strengthens the learning ability of IIDN with small computation overhead. Third, IIDN is superior to IIDN-non-attention in terms of both metrics. This proves that the attention mechanism helps the IIDN in understanding the user intent. Forth, the comparison between IIDN-non-attention and IIDN-non-auxiliary-task demonstrates that the auxiliary staying intent detection task exerts a stronger influence on the detection performance than the attention mechanism. Finally, IIDN outperforms IIDN-non-auxiliary-task by a significant margin about 4.38% and 22.83% in terms of AUC and Logloss. This verifies our conditional probability-based solution to the imbalance label problem for the purchasing intent detection task.

## 5.3 Allocation Evaluation (Online A/B Testing)

We deploy our allocation system in the real e-commerce environment and evaluate the allocation performance following the A/B testing methodology. The available amounts of coupons are 1 RMB, 2 RMB, 3 RMB and 5 RMB.

*5.3.1 Metrics.* At this stage, we focus on the performance of the coupon allocation (ie., the conversion rate and the cost). A metric named Increment Cost (IC) is used here to measure the cost to increase the number of buyers by one (Equation 11)

$$ic_\pi = \frac{C_\pi}{N_\pi(R_\pi - R_{\pi_0})} \quad (11)$$

where $\pi$ is the allocation policy, $C_\pi$ is the cost, $N_\pi$ is the number of users under the influence of $\pi$, $R_\pi$ is the conversion rate under the influence of $\pi$ and $R_{\pi_0}$ is the natural conversion rate (under the Non-allocation policy).

*5.3.2 Results.* Table 3 shows the performance of the compared allocation policies. All the allocation policies (except the Non-allocation policy) have the same budget constraint $B = 1M$ RMB. The increment cost of the Non-allocation policy is zero since no coupon coupon is allocated. The conversion rate under the non-allocation strategy is thus called the natural conversion rate. Generally speaking, the coupon has a positive effect on the conversion as the other strategies enjoy higher conversion rates. In particular, the

Table 3: Results of different coupon allocation strategies.

| Allocation Policy | Num. of Users | Num. of Buyers | Conversion Rate | Increment Cost |
|---|---|---|---|---|
| Non-Allocation | 1.22M | 130K | 10.7% | 0 |
| All-allocation | 1.21M | 166K | **13.7%** | 19.4 |
| Uplift-Allocation | 1.31M | 170K | 13.0% | 14.2 |
| IIDN-MCKP | 1.40M | 187K | 13.4% | **10.8** |

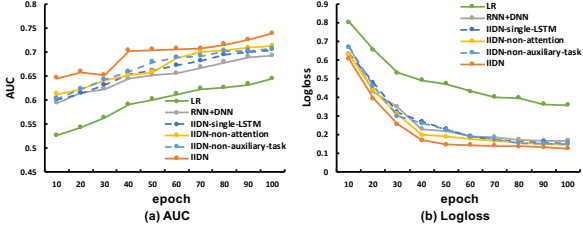

Figure 3: Learning curves with different components of IIDN.

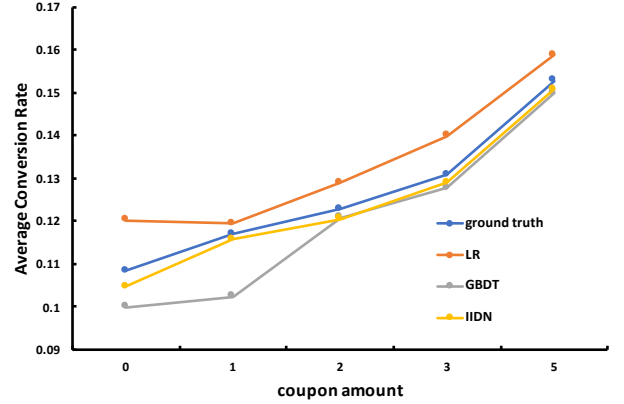

Figure 4: Monotonicity of different intent detection methods.

All-allocation strategy has the highest conversion rate. This is reasonable since all the users under this strategy are able to get the coupons. However, this strategy also suffers from the highest increment cost. In contrast, IIDN-Allocation contributes up to 25.69% conversion rate promotion with the lowest increment cost. The reason is that IIDN detects the user real-time intent precisely and the allocation system allocates the limited coupons to the proper user at the proper time.

### 5.4 Model Analysis

*5.4.1 Training Analysis.* In this section, we compare the effects of different components of IIDN on the training process. Figure 3 illustrates the training curves. We find out that IIDN converges more stable and faster than the other methods. The introduction of LSTM also speeds up the training process significantly. Moreover, although IIDN-non-attention achieves good results of AUC and Logloss, the learning speed at the early stage of the training process is relatively slow. This shows that the attention mechanism also accelerates the learning and helps extract key information from the input features.

*5.4.2 Monotonicity Analysis.* Normally, the greater the coupon amount is, the greater its stimulating effect is. That's logical because under the same conditions, users are more willing to place an order if more discounts are received. This hypothesis is also verified by the online results. Thus $pP(s, h, \cdot)$ should be a monotonically increasing function, which means a good intent detection model should hold this property. Figure 4 compares the monotonicity of the predicted purchasing probabilities using different models with different coupon amounts. The only non-monotonicity is brought by the LR model, which also suffers from the lowest AUC and the highest Logloss. Another observation is that predicting purchasing intents with small coupon amounts (RMB 0 or RMB 1) seems more difficult to models. This is because small amounts exert weak influence on users' final decisions. User intents will be harder to identify with these amounts. However, as the coupon amount increases, users are more likely to be stimulated by the coupon and the uncertainty of user intents is reduced.

*5.4.3 Budget Influence Analysis.* In our framework, the budget constraint is an important factor when we make the final decision. In this section, we study the influence of the budget constraint $B$ on the allocation performance. Table 4 presents the resulting conversion rate and increment cost under different budget cost. Two facts can be observed. (1) The conversion rate increases as the available budget increases. A slacker budget constraint means more users can receive the coupons and thus more users are stimulated. (2) As the budget amount increases, the increment cost also goes up. That's because under a slacker budget constraint, the allocation policy is less selective and some coupons are delivered to users who are not sensitive enough to coupons.

## 6 CONCLUSIONS

In this paper, we formulate the online coupon allocation problem. To overcome its challenges, we put forward a two-stage coupon allocation framework IIDN-MCKP. At the first stage, we detect the real-time user purchasing intent using IIDN. We propose several methods to overcome the obstacles at the detection stage. At the second stage, we model the allocation problem as a MCKP and coupons are allocated according to the intents predicted at the detection stage.

Extensive experiments show the superiority of our proposed framework on the coupon allocation task. The allocation system has made great profits for the platform and still functions online.

Table 4: Influence of the budget constraint on the conversion rate and the increment cost.

| Budget | Num. of Users | Num. of Buyers | Conversion Rate | Increment Cost |
|---|---|---|---|---|
| 0 | 1.23M | 133K | 10.8% | 0 |
| 50K | 1.23M | 134K | 10.9% | 5.93 |
| 100K | 1.23M | 137K | 11.1% | 6.13 |
| 500K | 1.23M | 140K | 11.3% | 8.24 |
| 1M | 1.23M | 167K | 13.5% | 10.5 |
| 5M | 1.23M | 179K | 14.4% | 33.8 |
| 10M | 1.23M | 200K | 16.2% | 55.6 |